\documentclass[conference]{IEEEtran}
\IEEEoverridecommandlockouts
\usepackage{cite}
\usepackage{amsmath,amssymb,amsfonts}
\usepackage{algorithmic}
\usepackage{graphicx}
\usepackage{textcomp}
\usepackage{xcolor}
\def\BibTeX{{\rm B\kern-.05em{\sc i\kern-.025em b}\kern-.08em
    T\kern-.1667em\lower.7ex\hbox{E}\kern-.125emX}}
\begin{document}

\title{Deep Learning for Classifying Food Waste}

\author{
\IEEEauthorblockN{Amin Mazloumian}
\IEEEauthorblockA{\textit{Institute of Embedded Systems} \\
\textit{Zurich University of Applied Sciences}\\
Zurich, Switzerland \\
amin.mazloumian@zhaw.ch}
\and
\IEEEauthorblockN{Matthias Rosenthal}
\IEEEauthorblockA{\textit{Institute of Embedded Systems} \\
	\textit{Zurich University of Applied Sciences}\\
	Zurich, Switzerland \\
	matthias.rosenthal@zhaw.ch}
\and
\IEEEauthorblockN{Hans Gelke}
\IEEEauthorblockA{\textit{Institute of Embedded Systems} \\
	\textit{Zurich University of Applied Sciences}\\
	Zurich, Switzerland \\
	hans.gelke@zhaw.ch}
}

\maketitle

\begin{abstract}
		One third of food produced in the world for human consumption -- approximately 1.3 billion tons -- is lost or wasted every year. By classifying food waste of individual consumers and raising awareness of the measures, avoidable food waste can be significantly reduced. In this research, we use deep learning to classify food waste in half a million images captured by cameras installed on top of food waste bins. 
		We specifically designed a deep neural network that classifies food waste for every time food waste is thrown in the waste bins. Our method presents how deep learning networks can be tailored to best learn from available training data.
\end{abstract}


\section{Introduction}
	A study carried out by the Swedish Institute for Food and Biotechnology, on request from the Food and Agriculture Organization of the United Nations revealed that roughly one third of the food produced in the world for human consumption is wasted or lost, which amounts to about $1.3$ billion tons per year \cite{gustavsson2011global}. In medium- and high-income countries about $40\%$ of food is wasted at the consumption stage \cite{gustavsson2011global,beretta2013quantifying}. To reduce waste at consumption stage, quantifying and classifying food waste is crucial \cite{reynolds2019consumption,secondi2015household,aschemann2015consumer}. However, there is a lack of data for individual consumers’ waste  \cite{reynolds2019consumption,secondi2015household}.
	
	In recent studies, relying on the paradigm of Internet-of-Things, waste data is monitored in smart garbage systems \cite{bhor2015smart,navghane2016iot}. In these smart systems, waste data is collected using sensors under waste bins. The collected sensor data is  periodically transferred to cloud for analysis and decision making. The sensors only measure waste level and waste weight in the bins.  These systems are primarily designed to accomplish timely waste collection. Interestingly however, by only providing food waste weight data to consumers and charging them accordingly, their food waste was reduced by $33\%$ \cite{hong2014iot}. 
	
	In this paper, a more informative view to food waste production behavior at the consumption stage is achieved through classifying food waste in waste bins. The classification task is feasible by processing images captured from food waste in the waste bins. The images are captured by installing cameras on top of the waste bins and monitoring the top surfaces of food waste in the bins. 
	This study focuses on classifying food waste in half a million images captured by cameras installed on top of waste bins. The system design of a smart garbage systems that uses our classification is out of the scope of this study.  
	
	The automatic classification of food waste in waste bins is technically a difficult computer vision task for the following reasons. a) It is visually hard to differentiate between edible and not-edible food waste. As an example consider distinguishing between eggs and empty egg shells. b) Same food classes come in a wide variety of textures and colors if cooked or processed. c) Liquid food waste, e.g. soups and stews, and soft food waste, e.g. chopped vegetables and salads, can largely hide and cover visual features of other food classes. 
	
	In this research, we adopt a deep convoultional neural network approach for classifying food waste in waste bins \cite{liu2017survey}. 
	Deep convolutional neural networks are supervised machine learning algorithms that are able to perform complicated tasks on images, videos, sound, text, and etc. The deep neural networks are composed of tens of convolutional layers (deep) that train on labeled data (supervised training) to learn target tasks. Labeled training data is composed of thousands of input-output pairs. In the training phase, the networks learn to produce the expected training output (labels) given the training input data. The training is performed by calculating millions of parameter values for feature extraction convolutional filters. In image processing, first layers of trained deep convolutional networks detect simple features, e.g. edges and corners. Based on the low level features extracted in first layers, deeper layers detect higher level features such as contours and shapes. 
	
	\emph{Image recognition convolutional networks} classify only one object of interest in every input image \cite{vgg16,inception,resnet}. For instance, a binary classifier trained to distinguish between cats and dogs can recognize whether its input image contains a cat or a dog. On the other hand, \emph{object detection convolutional networks} detect and locate multiple objects of interest in every image \cite{zhao2017survey}. The networks typically specify the location of every object  by defining a bounding box around the object. In our example, there would be a rectangle of type ``cat'' around every cat and a rectangle of type ``dog'' around every dog. Image segmentation networks classify every pixel in their input images \cite{guo2018review}. In the example, every pixel would be classified as a type cat pixel, a type dog pixel, or a background pixel. 
	
	Building training data for training object detection networks and segmentation networks are extremely costly. Thousands of images should be annotated by manually defining bounding boxes around every object when building training data for object detection networks. The situation is worse in building training data for segmentation networks. Every image contains tens of thousands of pixels, and every pixel should be marked with a class label.      
	
	What if there are multiple objects in every image, while localization and segmentation training data are not available? In our applications, multiple objects (food waste) are added sequentially to the monitored scenes (top surfaces of food waste in waste bins). In our available data set, for every time waste is thrown in the bins, only the class label of the waste is provided. 

	\section{Data}
	
	Our data set consisted of a total of about half a million images.
	The food waste bins were placed on weight sensors. Every time food waste was thrown inside a bin, the weight sensor under the bin detected the event and triggered an image capture. The images include the top surfaces of food waste, the inside parts of the waste bins, and very frequently surroundings of the waste bins. The surroundings were mainly parts of floor around the bin and very rarely plates and persons who threw the wastes. Each image received a time stamp together with a bin id, and a food class label. 
	
	The food classes of the images were manually labeled. There was a total number of $20$ food classes. Examples from food class labels are: apple, cheese, rice, and beef. The food class labels of images specify the last food class that was thrown in the bins. Food waste bins came in various shapes, sizes, colors, and materials. Some plastic and metal bins caused mirror reflections of food waste on the inside body of the bins. Another difficulty arose when garbage plastic bags were placed inside bins. In addition to the food class labels, for only $1000$ images, binary masks were provided that marked food waste pixels in the images. 
	
	\section{Method}
	
	Our proposed method has two sequential parts: a preprocessing pipeline and a classification convolutional network. The preprocessing pipeline prepares the images for the following classification part. The pipleine performs scaling, background subtraction, and region-of-interest cropping. Such processes are typical in image processing applications. However, we specifically designed a deep neural network to well benefit from our available training data.
	
	\subsection{Preprocessing Pipeline}\label{sec:preprocessing}

	\begin{figure*}
		\centering
		\includegraphics[width=0.85\textwidth]{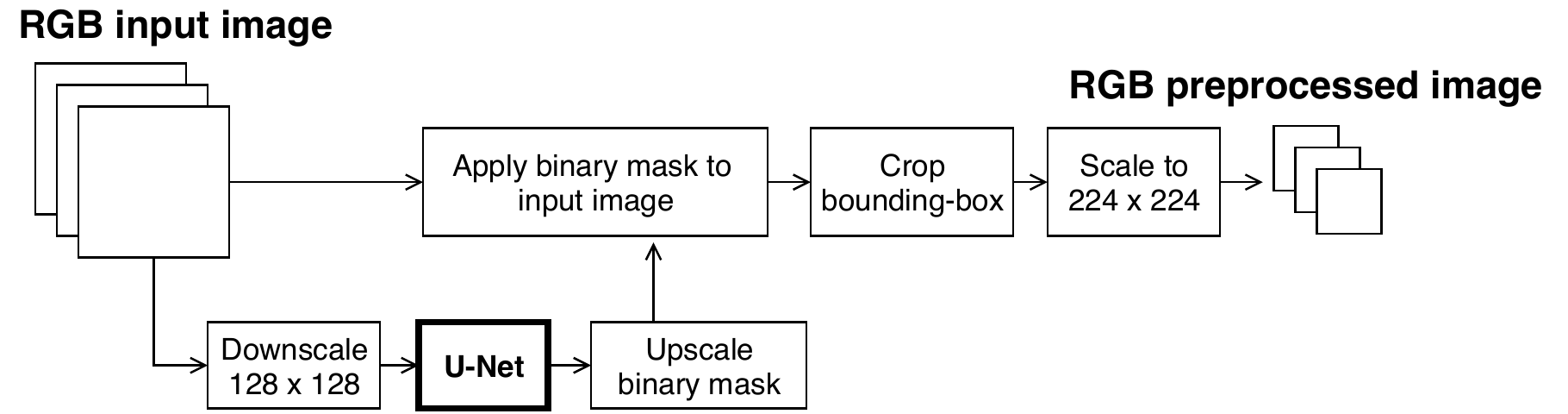}
		\caption{In a preprocessing pipeline, food waste pixels in captured images are marked using a U-Net. Square-shaped bounding boxes for the marked pixels are then automatically calculated. Finally, the bounding boxes are cropped and scaled to a desired fixed size for classification.}\label{preprocessing}
	\end{figure*}

	We trained a U-Net deep convoultional network for background subtraction \cite{unet2015}. More specifically, we used the $1000$ mentioned images with binary masks to train a deep convolutional network with the U-Net architecture to mark food waste pixels in the captured images. The U-Net architecture relies on the strong use of data augmentation to use available annotated samples very efficiently. It is therefore feasible to train a U-Net on the very few available annotated images for which binary masks are available.

	In our application, we used a U-Net the binary classification of pixels for one of the two classes: a) food waste and b) not food waste. By only selecting food waste pixels, the food waste classification in our next step would not have to deal with 
	the other parts of the images.
	Our trained U-Net produces food waste binary masks of size $128 \times 128$ for an input RGB image of the same resolution. The network includes about $8.6$ million trainable parameters and $19$ convoultional layers.
	
	The complete preprocessing pipeline is illustrated in Figure \ref{preprocessing}. After extracting food waste pixels using the trained U-Net, a square-shaped bounding box of detected food waste is calculated and cropped. The bounding box squares are squares with minimum sizes that include all food waste pixels within images. The cropped bounding boxes are finally scaled to a desired fixed size ($224 \times 224$) for the classification in the next step.
	
	\subsection{Deep Neural Network for Classifying Food Waste}\label{sec:deep}
	The classification labels only specify the food classes of the waste that were \emph{last} thrown in the waste bins. As an example, imagine that we  throw some salad in an empty waste bin. The added weight of the waste bin would be detected by the weight sensor under the bin. This would trigger an image captured from the top surface of salad. The image would be later manually annotated as food class salad. Now imagine that we throw an apple in the same waste bin. Another image would be captured and would be annotated as food class apple. The new image would not only contain the apple but also the salad around the apple. Parts of apple might be covered by the  salad and the surface appearance of salad would change around the apple. As we throw more food waste is the bin, more food classes would be present in the captured images. Although the top surface appearance of previously-existing waste in the bins slightly changes every time, the label specifies only the food waste class of the last thrown waste.   
	
	To classify the last food waste thrown in the bins, it should be identified what parts of images belong to the last food waste thrown in the bins. Therefore, every time the images of the waste bin \emph{before} and \emph{after} throwing  waste should be compared. As an image is captured for every time waste is thrown in the bins, consecutive pairs of images from same bins should be compared. It is naive to assume that pixel values that have significantly changed between the consecutive images necessarily belong to the last thrown food waste. Because, once food waste is thrown in the bins, the pixels values belonging to the visible surface of the previously-existing food waste potentially  changes as well.
	
\begin{figure*}
		\centering
		\includegraphics[width=0.95\textwidth]{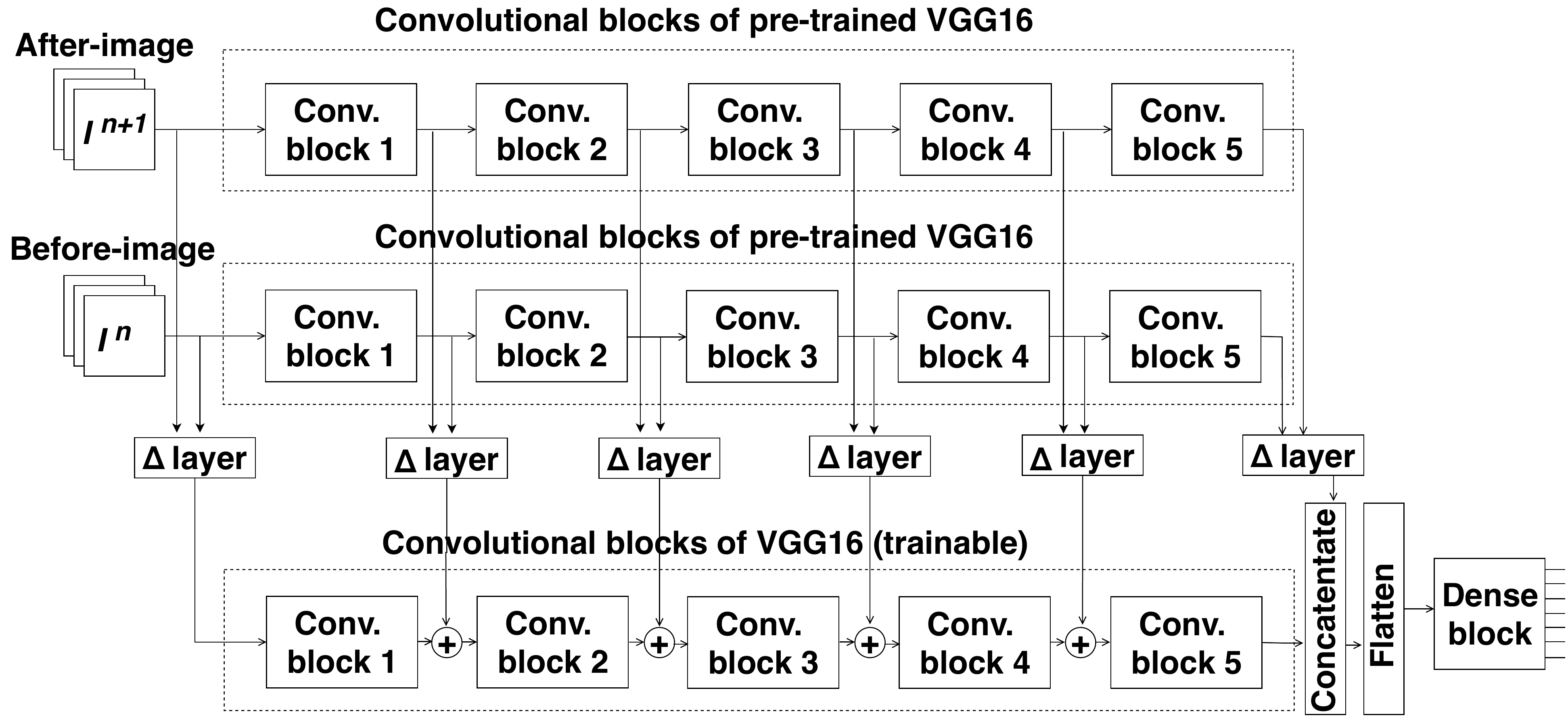}
		\caption{Our deep neural network consists of two seperate pre-trained VGG16-based convolutional paths for feature extraction in the images captured before and after throwing food waste. The feature volumes of the two convolutional paths at different abstraction layers are integrated in a third trainable convolutional path using our proposed delta layers.}\label{deltanet}
	\end{figure*}

	We designed a deep convolutional neural network that classifies last food waste thrown in the bins based on the two images before and after throwing the waste (Figure \ref{deltanet}). The features of the two images are extracted at different abstraction layers through two separate convolutional paths. While initial convolutional layers extract low-level features such as edges and corners, deeper convolutional layers extract highter-level features such as contours and shapes at higher abstraction levels. The two covolutional paths are two instances of the convolutional blocks of a pre-trained VGG16 trained on the Imagenet dataset \cite{vgg16}. During the training phase, we set the weights of these two feature extraction paths non-trainable (freeze the two convolutional paths). The VGG16 consists of $5$ convolutional blocks. Each convolutional block contains several convolution layers and a pooling layer. The pooling layers down-sample the representation of the image. The first two convolutional blocks of a VGG16 network contain $2$ convolutional layers, while the last three convolutional blocks contain $3$ convolutional layers.
	
	In our design, we introduce a new neural network layer, \emph{delta layer}, to emphasize features that appear after throwing the last food waste. After throwing an apple on salad, for example, we aim to focus on location-specific features of the apple-and-salad image, which were not present in the salad image. The delta layer at a certain abstraction layer computers $\psi\left(V_{\text{after}}-\lambda^c V_{\text{before}}\right)$ for feature volumes of the after-image $V_{\text{after}}$ and before-image $V_{\text{before}}$, where $\lambda^c$ is a trainbale scaling constant and $\psi$ denotes the ReLU activation function. The ReLU activation function sets negative values to zero and keeps positive values unchanged. The intuition behind our layer design is to keep feature values that exist (have big positive values) in $V_{\text{after}}$ and do not exist (have smaller positive values or negative values) in $V_{\text{after}}$.   
	
	The feature volumes of the two convolutional paths at different abstraction layers are integrated in a third \emph{trainable} convolutional VGG16 path using delta layers. More specifically, the resulting  feature volumes of delta blocks at each abstraction level are added to result of the same abstraction layer of the third convolutional path. Finally, the last feature volume of the third path is flattened and after three fully-connected neural network layers (dense block), classification output is produced. Our deep model has in total $50.7$ million parameters from which $21.2$ million are trainable and $29.5$ million are not trainable.
	
	\section{Results}
	
	In a pre-processing phase, we used a U-Net for background subtraction (Section \ref{sec:preprocessing}). For $1000$ images segmentation binary masks were available. The masks marked food waste pixels in the images. We split the data set into training, validation, and test sets with $700$, $200$, and $100$ images, respectively. The U-Net was trained to a pixel-wise accuracy of $95.8\%$. In agreement with the literature, we observed that with a relatively small data set our U-Net was well trained.  
			
	To classify food waste, we designed a deep convolutional neural network (Section \ref{sec:deep}). The classification data set consisted of about half a million images. Each image was labeled only with a food class label. The label defines the food waste class of the last thrown item in the waste bin. We emphasize that although multiple food waste classes were available in images, no localization and segmentation data were available. Our trained deep network, classifies food waste images into $20$ food waste classes with a categorical accuracy of $83.5\%$.

	\begin{figure*}
		\centering
		\includegraphics[width=0.85\textwidth]{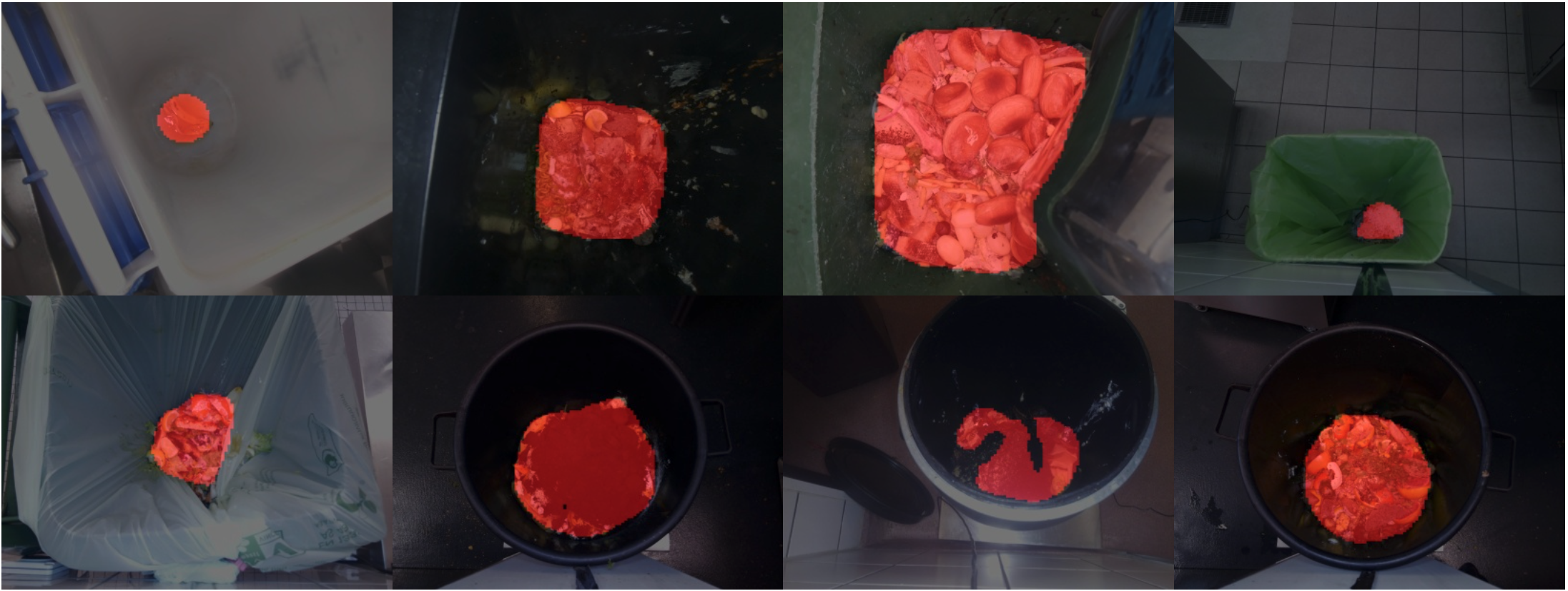}
		\caption{In a pre-processing phase, a trained U-Net network detects for every pixel whether is belongs to food waste. The pixel-wise accuracy of the trained U-Net is $95.8\%$. In this figure, food waste pixels are marked with a red tone.}\label{foodwaste}
	\end{figure*}

	\section{Discussion}
In this study, we adopted a deep convoultional neural network approach for classifying food waste in waste bins. In this study, multiple objects of interest (food wastes) were added sequentially to a monitored scene (food waste bin). As multiple objects of interest were present in every image, image recognition networks were not applicable. Since localization data was not provided, we could not use object detection networks. Also, segmentation data was not available. We therefore could not use segmentation networks. 

In our data set, each image had a single label. The classification labels only specified the food class that was last thrown in the waste bins. We designed a deep convolutional neural network that classified last food waste thrown in the bins based on the two images before and after depositing the waste. The features of the two images were extracted at different abstraction layers through two separate convolutional paths. The feature of the two convolutional paths at different abstraction layers were integrated in a third convolutional path using our specifically-designed delta layers. Finally, the last feature volume of the third path was flattened and after three fully-connected neural network layers (dense block), classification output was produced. Our trained deep network classified food waste images into $20$ food waste classes with a categorical accuracy of $83.5\%$. Our results show how deep learning networks can be tailored to best learn from available training data.

\section*{Acknowledgment} This research was funded through an Innosuisse grant by the Swiss Innovation Agency (2019-2020).

\bibliographystyle{IEEEtran}
\bibliography{conference}

\end{document}